\documentclass[twoside,leqno,twocolumn]{article}

\usepackage[letterpaper]{geometry}

\usepackage{ltexpprt}
\usepackage{hyperref}
\usepackage{booktabs}
\usepackage{graphicx}
\usepackage{amsmath}
\usepackage{multirow}

\usepackage[maxbibnames = 1, sorting = none]{biblatex} 
\begin{document}

\newcommand\relatedversion{}
\renewcommand\relatedversion{
} 

\title{\Large Multiple Relations Classification using Imbalanced Predictions Adaptation\relatedversion}
\author{Sakher Khalil Alqaaidi\thanks{School of Computing, University of Georgia. \textit {\{sakher.a, elika.bozorgi, kkochut\}@uga.edu}
}
\and Elika Bozorgi\footnotemark[1]
\and Krzysztof J. Kochut\footnotemark[1]
}

\date{}

\maketitle


\fancyfoot[R]{\scriptsize{Copyright \textcopyright\ 2024 by SIAM\\
Unauthorized reproduction of this article is prohibited}}





\begin{abstract} \small\baselineskip=9pt The relation classification task assigns the proper semantic relation to a pair of subject and object entities; the task plays a crucial role in various text mining applications, such as knowledge graph construction and entities interaction discovery in biomedical text. Current relation classification models employ additional procedures to identify multiple relations in a single sentence. Furthermore, they overlook the imbalanced predictions pattern. The pattern arises from the presence of a few valid relations that need positive labeling in a relatively large predefined relations set. We propose a multiple relations classification model that tackles these issues through a customized output architecture and by exploiting additional input features. Our findings suggest that handling the imbalanced predictions leads to significant improvements, even on a modest training design. The results demonstrate superiority performance on benchmark datasets commonly used in relation classification. To the best of our knowledge, this work is the first that recognizes the imbalanced predictions within the relation classification task.\end{abstract}

\section{Introduction} \label{sec:intro} The relation classification (RC) task aims to identify relations that capture the dependency in every pair of entities within unstructured text. The task is employed in several applications, such as knowledge graph construction and completion \cite{chen2020knowledge} and entities interaction detection in biomedical text \cite{bundschus2008extraction}. In knowledge graphs, it is common to employ relational triples as the base structure. A triple consists of a subject entity, an object entity, and a semantic relation connecting them. For instance, Wikipedia articles rely on Wikidata knowledge base to provide its content \cite{vrandevcic2014wikidata}; users can query Wikidata in a structured format using SPARQL and retrieve the information as RDF triples. In biomedical text, the RC task helps in discovering the interactions between entities such as proteins, drugs, chemicals and diseases in medical corpora.

In the supervised RC task, the objective is to learn a function that takes a sentence and its tagged entities as input, then assigns a binary class to each relation from a predefined set. A positive label indicates that the relation is valid for an entity pair. Thus, the output consists of the positive relations. We use this formal notation for the task:

\begin{equation}
    \label{eq:task}
    f(W, E, P) = \left\{ \begin{array}{ll}
         R, & \textrm{Multiple relations} \\
         r, & \textrm{Single relation} \\
         \emptyset, & \textrm{otherwise}
    \end{array}
    \right.
    \end{equation}
where $W$ is a sequence of words $[w_1,\ w_2\ ...\ w_n]$, $E$ is the set of one or more entity pairs. Each entity pair consists of a subject entity and an object entity, where an entity is a sub-sequence of $W$. $P$ is the predefined relations set. $R$ is a set of multiple relations found for $E$. $r$ is a single relation. $\emptyset$ indicates that no relation exists connecting any of the entities. In an example from the Nyt dataset \cite{riedel2010modeling} with the sentence \textit{``Johnnie Bryan Hunt was born on Feb. 28 , 1927 , in rural Heber Springs , in north-central Arkansas.''}, the valid relations are \textit{``contains"} and \textit{``place lived"} for the entity pairs \textit{(``Arkansas", ``Heber Springs")} and \textit{(``Johnnie Bryan Hunt", ``Heber Springs")}, respectively.

Usually, a sentence incorporates multiple relations. Table \ref{tab:relations} shows the average number of relations in two well used benchmarks \cite{riedel2010modeling,zeng2018extracting}. Therefore, a single RC approach is only valid for limited cases. However, majority of the literature work follow the single relation approach. Single RC models require additional preprocessing procedure to be able to identify multiple relations \cite{wang-etal-2019-extracting}, that is by replicating the sentence $W$ in equation \ref{eq:task}, then assigning an entity pair and a single relation $r$ to each copy. Such approach does not only incur additional steps but also an added training load. An additional downside is losing the contextual information due to splitting the entities data in the input \cite{qu2014senti, yin2006efficient}, which would result missed accuracy enhancements. Besides that, several single RC models evaluate their work on highly class-imbalanced benchmarks, such as Tacred \cite{zhang2017tacred} or datasets with a few predefined relations. For instance, SemEval \cite{hendrickx-etal-2010-semeval} has only six relations. Such performance measurements make it hard to generalize to real-world scenarios. Additionally, these models employ complicated approaches, such as attention mechanisms, additional training and tuning efforts \cite{wang2016relation, zhou2016attention}. Furthermore, most approaches neglect the imbalanced prediction pattern in the predefined relations set, when the model learns to predict only one relation out of many others in the predefined set.

The multiple RC approach tackles the previously mentioned problems. However, regular methods still unable to achieve competitive results, mainly affected by the need to adapt to the imbalanced prediction. Despite the ability to predict several relations, their number is relatively smaller than the predefined relations set. This gap is shown in Table \ref{tab:relations} when comparing the average number of relations with the predefined set size, which indicates high imbalanced distribution of positive and negative labels in each sentence. Furthermore, the table shows the percentage of sentences of three or more prediction relations, reflecting the importance of the multiple RC task.

\begin{table}
\centering
\caption{The number of predefined relations in the Nyt and Webnlg datasets, the average number of positive relations in each sentence, the standard deviation, and the percentage of sentences with 3 or more positive relations.}
\label{tab:relations}
\begin{tabular}{lcccc}
\toprule
Dataset & Relations & Avg. & Stdev. & 3+ Rels.\\
\midrule
Nyt & 24 & 2.00 & 2.88 & 18.48\%\\
Webnlg & 216 & 2.74 & 2.23 & 41.72\%\\
\bottomrule
\end{tabular}
\end{table}

In this paper, we propose a \textbf{M}ultiple \textbf{R}elations \textbf{C}lassification model using Imbalanced Predictions \textbf{A}daptation (MRCA). Our approach adapts to the imbalanced predictions issue through adjusting both the output activation function and the loss function. The utilized loss function has proved its efficiency in several imbalanced tasks. However, our customization shows additional enhancements within the RC task. Furthermore, we utilize the entity features through concatenating an additional vector to the word embeddings in the text encoder level.

The evaluation shows that our approach outperforms other models that reported their multiple RC performances in the relation extraction task on two popular benchmarks. To the best of our knowledge, this is the first work that addresses the imbalanced predictions within the RC task. The ablation study demonstrates the efficacy of our approach components in adapting to the imbalanced predictions, and in utilizing the text and the entity features. Furthermore, the architecture of our model has a light design that yields astonishing performance. We make our code available online\footnote{https://github.com/sa5r/MRCA}.

\section{Related Work} \label{sec:related}

\subsection{Single Relation Classification}
Generally, RC models pursued efficient text representation to identify relations. Early supervised approaches \cite{wang2008re, fundel2007relex} employed natural language processing (NLP) tools to extract text features, such as word lexical features, using dependency tree parsers \cite{klein2002fast}, part-of-speech (POS) taggers and named entity recognition. Relex \cite{fundel2007relex} generated dependency parse trees and transformed them into features for a rule-based method.

With the achievements of neural network methods, deep learning models utilized a combination of text lexical features and word embeddings for the input \cite{gormley2015improved, zhang2018graph} while other approaches \cite{zhou2016attention, zeng2014relation, lee2019semantic,ding2022relation} depended on those embeddings solely to avoid NLP tools error propagation to later stages \cite{zeng2014relation}. Neural network-based models employed word embeddings in different ways. First, embeddings generated from algorithms such as Word2Vec \cite{mikolov2013efficient} using custom training data such as in \cite{gormley2015improved, zeng2014relation}. Second, embeddings from pre-trained language models (PLMs), such as Glove \cite{pennington2014glove}. These PLMs were utilized in the works including \cite{zhou2016attention, zhang2018graph, lee2019semantic, ding2022relation}. In \cite{zhou2016attention}, authors presented a neural attention mechanism with bidirectional LSTM layers without any external NLP tools. In C-GCN \cite{zhang2018graph}, the dependency parser features were embedded into a graph convolution neural network for RC. TANL \cite{paolini2021structured} is a framework to solve several structure prediction tasks in a unified way, including RC. The authors showed that classifiers cannot benefit from extra latent knowledge in PLMs, and run their experiments on the T5 language model.

Bert \cite{devlin2018bert} is a contextualized PLM that has presented significant results in various NLP tasks and several RC models employed it \cite{wu2019enriching, baldini-soares-etal-2019-matching, cohen2020relation, 9894216}. The earliest was R-Bert \cite{wu2019enriching}, where authors customized Bert for the RC task by adding special tokens for the entity pairs. Later, Bert’s output was used as an input for a multi-layer neural network. In \cite{cohen2020relation}, the traditional classification was replaced with a span prediction approach, adopted from the question-answering task. In \cite{9894216}, the model combined short dependency path representation generated from dependency parsers with R-Bert generated embeddings.

\begin{figure*}[h!]
  \includegraphics[width=\textwidth]{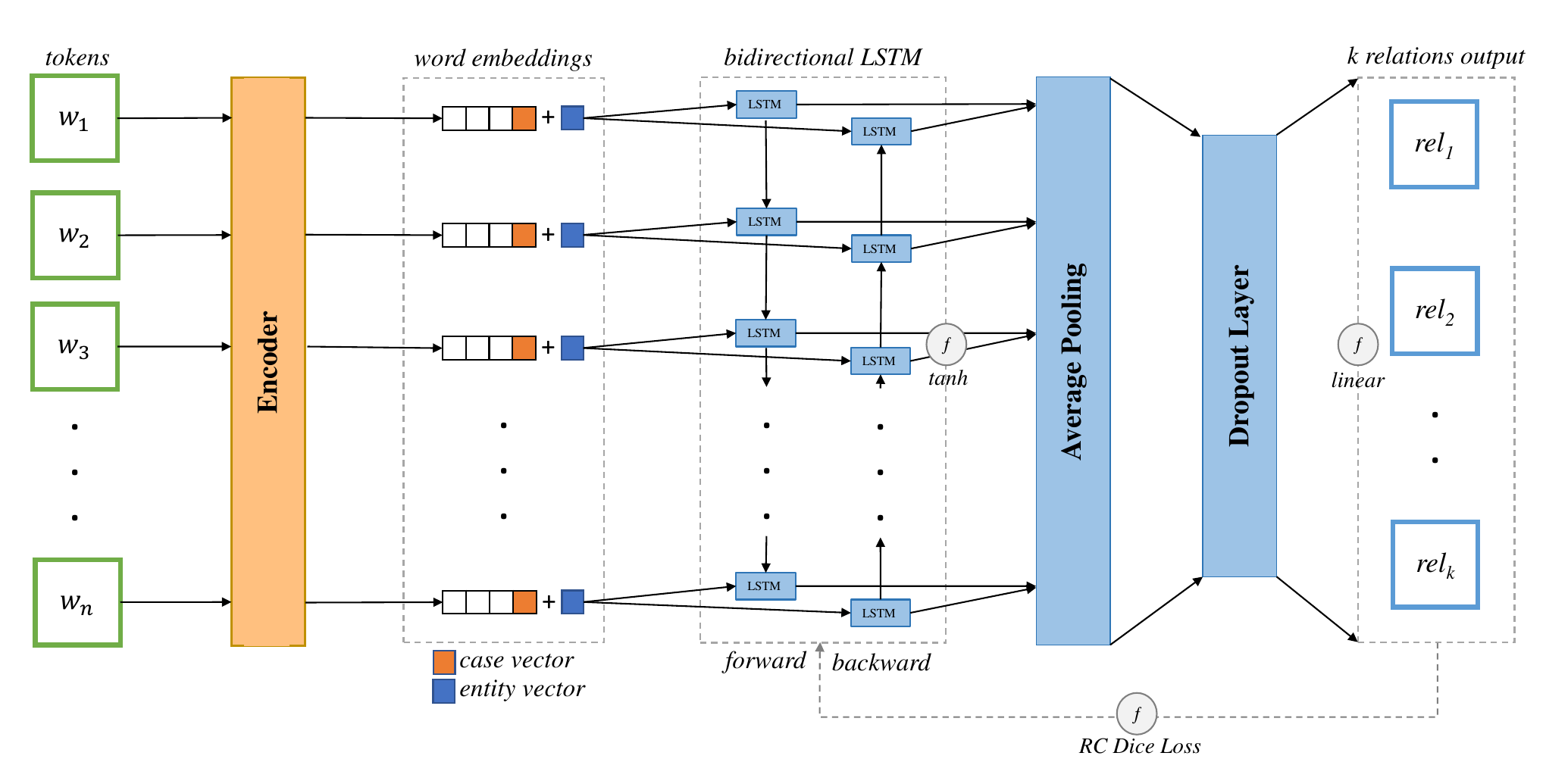}
  \caption{The main architecture of our model. The adaptation approach uses a linear activation function in the output and the Dice loss extension. Furthermore, we enhance the embeddings by adding two vectors, a character case vector and an entity type vector denoted by the orange and blue squares.}
  \label{fig:diagram}
\end{figure*}
\subsection{Multiple Relations Classification} Methods that classify multiple relations in a single input pass vary based on the usage of NLP tools, neural networks and PLM models. Senti-LSSVM \cite{qu2014senti} is an SVM-based model that explained the consequences on the performance when handling multi-relational sentences using a single relation approach.

CopyRE \cite{zeng2018extracting} is an end2end entity tagging and RC model that leveraged the copy mechanism \cite{gu2016incorporating} and did not use a PLM. Instead the model used the training platform’s layer to generate word embeddings. In the RC part of the model, the authors used a single layer to make predictions over the softmax function. Inspired by CopyRE, CopyMTL \cite{zeng2020copymtl} is a joint entity and relation extraction model with a seq2seq architecture. The model followed CopyRE's approach in representing text.

Several models employed Bert in the RC task \cite{wang-etal-2019-extracting,li2020downstream}. The work in \cite{wang-etal-2019-extracting} elaborated on the flaws of the single relation prediction in multi-relational sentences and presented a model that is based on customizing Bert. Specifically, the model employed an additional prediction layer and considered the positions of the entities in the input. In \cite{li2020downstream}, authors showed that RC is not one of the training objectives in the popular PLMs. Therefore, they leveraged Bert and used a product matrix to relate the identified relations to the sentence entities.

GAME model \cite{cheng2022multi} used the NLP tool Spacy \cite{honnibal2017spacy} to generate word embeddings. The model is based on graph convolution networks for global sentence dependency and entities interaction features. ZSLRC \cite{gong2021zero} is a zero-shot learning model that used Glove PLM. We mention this work because it reports the supervised learning performance in RC task.

\section{Methodology}
Our model incorporates two components, an output adaptation module and an input utilization techniques. Between the two implementations, we employ a light design to achieve low training parameters and better performance. We use an average pooling layer to reduce the dimensionality of the network before the output layer. The dropout layer is used to tackle training overfitting. Finally, in the output layer, each unit represents a relation. Figure \ref{fig:diagram} shows the main architecture of our model.

\subsection{Text Encoder} \label{sec:encoder}
We utilize Glove \cite{pennington2014glove} pre-computed word embeddings to encode the input sentences. Glove embeddings are retrieved from a key-value store where words in lowercase are the keys for a float vectors matrix $R^{s \times d}$, where $s$ is the vocabulary size and $d$ is the embedding dimensions. We find Glove more convenient for the task to tackle the out-of-vocabulary (OOV) \cite{woodland2000effects} problem. Specifically, Glove's most used variant\footnote{https://nlp.stanford.edu/projects/glove/} has relatively large dictionary of 400,000 words. However, the embeddings are context-free and the keys are case insensitive. Other popular PLMs have much smaller vocabularies but support Glove's missed features. For instance, Bert \cite{devlin2018bert} generates contextual embeddings and has character case support. Nevertheless, the commonly used Bert variant\footnote{https://tfhub.dev/tensorflow/bert\_en\_uncased\_L-12\_H-768\_A-12/4} has 28,997 vocabulary entries only. Thus, OOV words will get its representation based on the latent training parameters \cite{nayak2020domain}. At the same time, several studies showed that RC is not one of the training objectives in Bert \cite{li2020downstream, liu2019roberta}. Thus, we adjust Glove to provide the missed features as the following.

First, having case sensitive embeddings is essential to denote entity words in the sentence. Realizing entities in the RC task is crucial to detect the proper relation. Generally, a word with an uppercase first character is an entity word. Thus, we add an additional vector to the word embeddings to denote the first character case. For uppercase first character words we use the value of ceiling the largest vector value in Glove. Formally, the vector value is computed as the following:

\begin{equation}
    \label{eq2}
    v = \lceil \max_{1\le i \le s}{( \max_{1\le j \le d}{ ( R[i][j] ) } )} \rceil
\end{equation}
where $R$ is the vectors matrix in Glove, $s$ is the vocabulary size, and $d$ is the embedding dimensions. For lowercase first character words, we use the negative value of $v$. We employ the maximum and minimum values in the PLM to boost distinguishing entity words from non-entity words. The orange square in Figure \ref{fig:diagram} denotes the first character case vector.

Second, to provide contextual sentence representation, we make us of a bidirectional long short-term memory (LSTM) as our first layer in the model architecture.

Although we employ large vocabulary in encoding the sentence, a few words are still not matched. Thus, generate their embeddings by combining the character level embeddings.

\textbf{Entity Features} We show in equation \ref{eq:task}, that the task input consists of subject and object entities in addition to the sentence. We attempt to enrich the input with these details by following a similar approach of appending an additional vector from section \ref{sec:encoder}. Specifically, we append a vector of the value $v$ from equation \ref{eq2} to the word representation when the input indicates that the word is a subject entity or part of it, the negative value of $v$ for object entity words, and 0 for non-entity words. The dense blue square in Figure \ref{fig:diagram} denotes this additional vector. Formally the vector is given by the function $f_{entVec}$ as the following:

\begin{equation}
\label{eq1}
f_{entVec}(w) = \left \{ \begin{array}{lll}
v&,&w \in E_{sub}\\
-1\times v&,&w\in E_{obj}\\
0&,&w\notin \{ E_{sub}\cup E_{obj}\}
\end{array}
\right.    
\end{equation}
where $w$ is a word in the sentence, $E_{sub}$ is the subject entities set and $E_{obj}$ is the object entities set. We use the negative value in the object entity to emphasize the difference between entity types and make the relation direction between entity pairs recognizable while training.

\begin{figure}[h]
  \includegraphics[width=0.48\textwidth]{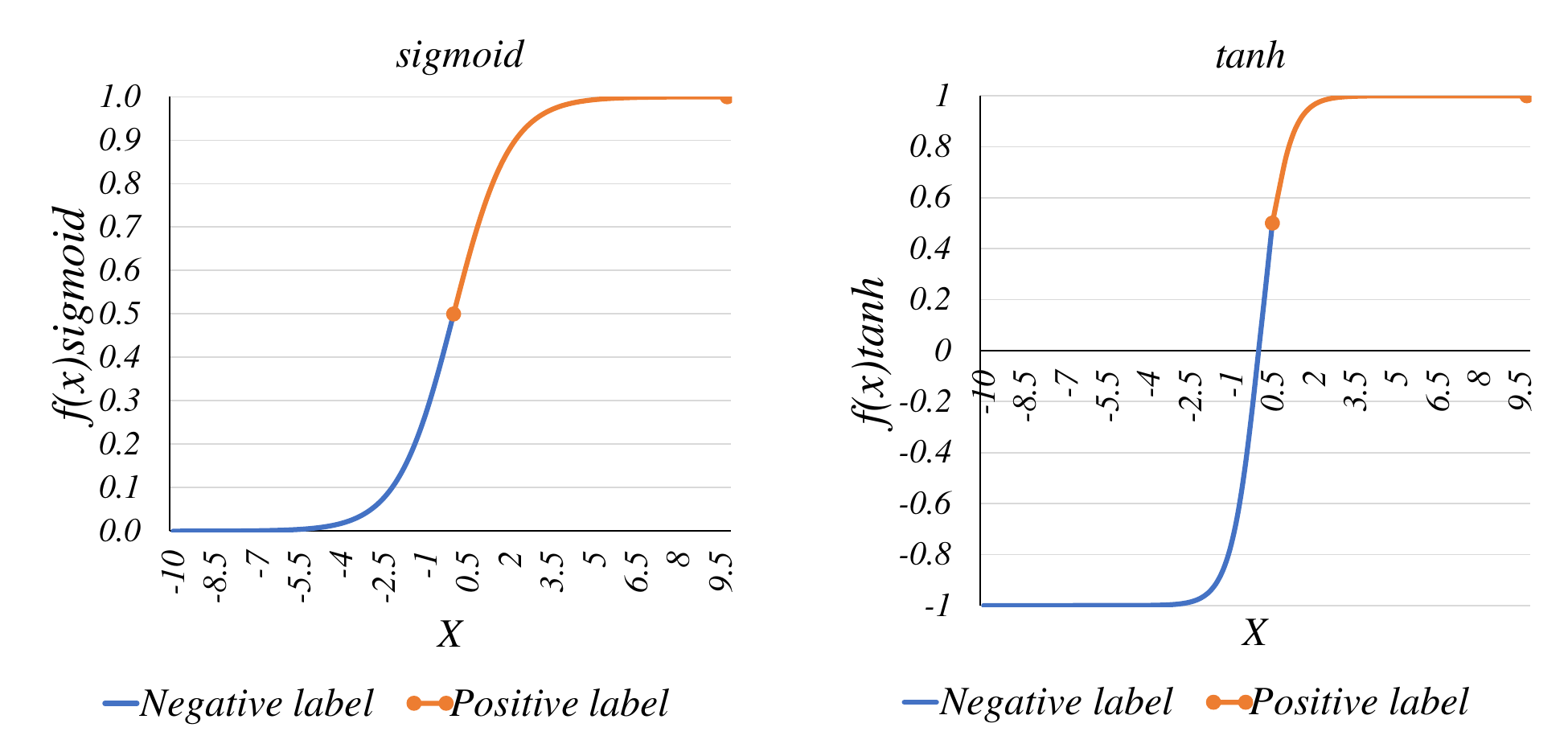}
  \caption{Comparison between prediction ranges in the sigmoid function and our implementation}
  \label{fig:sigtanh}
\end{figure}
\subsection{Imbalanced Predictions Adaptation} \label{sec:adaptations}
In real-world scenarios, the number of predefined relations is usually greater than the number of positive relations in a single sentence by a big ratio. Consider the gap in Table \ref{tab:relations} between Webnlg relations and the average number of valid relations in each sentence. We see that it is impractical to employ traditional probability activation functions in neural networks (NN) for this case. For instance, \textit{sigmoid} and \textit{softmax} are commonly used functions in NNs \cite{chollet2021deep}. Our claim is supported by the fact that these functions treat positive and negative predictions equally. In other words, all probability predictions of 0.5 or greater are considered as positive label predictions in the mentioned functions. Thus, we improve the model's ability to predict negative labels by devoting 75\% of the prediction range for the negative labels. We implement this step by restricting the model's layers output to a value within the range of -1 and 1. We perform that through applying \textit{tanh} activation function to the first layer, then using a linear activation function in the output layer. As a result, three quarters of the range are used for the negative labels, i.e., all predictions between -1 and 0.5 indicate a negative label. Figure \ref{fig:sigtanh} compares the prediction ranges in a probability activation function (\textit{sigmoid}) and the output of the \textit{tanh} activation function.

\textbf{Dice Loss Extension} Traditionally, straightforward classification models employ the cross-entropy loss functions \cite{chollet2021deep}, that are used to improve the accuracy, whereas the RC task objective is to reduce the false positive and false negative predictions. Thus, we seek improving the precision and recall performances, i.e., enhancing the model’s f1 score. Dice Loss has shown significant results in several domains, such as computer vision \cite{huang2018robust} and other NLP tasks that have imbalanced data \cite{li-etal-2020-dice}. The function was designed with inspiration of the f1 metric as the following:
\begin{equation}
DiceLoss(y_i, p_i) =  1 - \frac{2p_iy_i + \gamma}{ p_i^2 + y_i^2 + \gamma}
\end{equation}
where $y_i$ is the ground-truth label for relation $i$, $p_i$ is the prediction value, and $\gamma$ is added to the nominator and the denominator for smoothing, which has a small value of 1e-6 in our implementation.

\begin{table}[t]
\centering
\caption{Loss calculations for ground truth $y$ and the prediction value $p$ in Dice loss and in our implementation. The underlined numbers are the unconventional values in Dice loss  }
\label{tab:dice}
\begin{tabular}{lcccc}
\toprule
$y$&$p$&Expected loss&Dice loss&RC Dice loss\\
\midrule
0 & 1 & $\geq$ 1 & 0.9 & 0.9 \\
0 & 0.1 & $\approx$ 0 & \underline{0.9} & 9e-13 \\
0 & -0.1 & $\approx$ 0 & \underline{0.9} & 9e-13 \\
0 & -1 & 0 & \underline{0.9} & 9e-13 \\
1 & 1 & 0 & 0 & 0 \\
1 & 0 & $\geq$ 1 & 0.9 & 0.9 \\
1 & -1 & \textgreater 1 & 1.9 & 1.9 \\
\bottomrule
\end{tabular}
\end{table}

Utilizing  Dice Loss in our adapted predictions may incur unconventional behaviour. Specifically, when having negative ground truth labels and negative value predictions at the same time. Such case would result high loss when using Dice Loss, whereas low loss is the natural result. Our analysis in Table \ref{tab:dice} shows the invalid loss values and the expected ones. Therefore, we expand our adaptation by implementing an extension for Dice Loss. Specifically, we address the negative prediction case by computing the loss from a division operation; the nominator is the squared smoothing value; the denominator is the regular Dice loss denominator. Raising the smoothing value to the second power is necessary to present a small loss value. Our corrected loss value examples can be observed in Table \ref{tab:dice}. We call this extension, \textit{RC\_DiceLoss} and formally define as the following:
\begin{equation}
\begin{split}
&RC\_DiceLoss(y_i, p_i)=\\
&\left \{ \begin{array}{lll}
\frac{\gamma^2}{ p_i^2 + y_i^2 + \gamma}&,&y_i = 0 \: and \: p_i \textless 0.5\\
\\
1 - \frac{2p_iy_i + \gamma}{ p_i^2 + y_i^2 + \gamma}&,&otherwise\\
\end{array}
\right.
\end{split}
\end{equation}

\section{Experiments}
\subsection{Datasets and Experimental Setup} To demonstrate the generalization and the applicability of our model, we evaluated it on diverse and widely used datasets. The Nyt dataset \cite{riedel2010modeling} was generated from a large New York Times articles corpus, where each input item consisted of a sentence and a set of triples, each triple is composed of subject and object entities, and a relation. Webnlg dataset was originally generated for the Natural Language Generation (NLG) task, CopyRE \cite{zeng2018extracting} customized the dataset for the triples and relations extraction tasks. Table \ref{tab:datasets} shows the statistics and the splits of the datasets.

Our model achieved the best results using Glove PLM. The language model has been trained on 6 Billion tokens with a 400,000 words vocabulary and 300 dimensional word embeddings. Nevertheless, the experiments demonstrated that our model can adopt other PLMs and still provide competitive results. We performed the experiments using TensorFlow. Our model's hyper-parameters and training settings are the unified for both experimental datasets, which confirms the applicability of our approach to real-world data. Table \ref{tab:tuning} shows the training settings and the model hyper-parameters. We used Adam optimizer for stochastic gradient descent, and performed the training for five times on every dataset with different random seed and reported the mean performance and the standard deviation. Although we implement the training for 50 epochs, the mean convergence epoch for the Nyt dataset was 21.4. The hyper-parameters were chosen based on tuning the model for best performance. We ran the experiments on a server with NVIDIA A100-SXM-80GB GPU device and AMD EPYC MILAN (3rd gen) processor, but using only 8 cores. We used only 20GB of the available main memory for the Webnlg dataset experiments and 100GB for the Nyt dataset due to its size. We conducted an ablation study to test our model’s components using different variants as shown in Section \ref{sec:ablation}.

\begin{table}[t]
\centering
  \caption{Statistics of the evaluation datasets.}
  \label{tab:datasets}
  \begin{tabular}{llcc}
    \toprule
    Dataset& &Nyt&Webnlg\\
    \midrule
    Relations & & 24 & 216\\
    \hline
    \multirow{4}{*}{Samples} & Training  & 56,196 & 5,019\\
    &Validation & 5,000 & 500\\
    &Testing & 5,000 & 703\\
    \cline{2-4}
    &Total & 66,196 & 6,222\\
  \bottomrule
\end{tabular}
\end{table}

\begin{table}[t]
\centering
  \caption{Model hyperparameters and training settings.}
  \label{tab:tuning}
  \begin{tabular}{llc}
    \toprule
     Parameter & & Value\\
    \midrule
    \multirow{2}{*}{Average Pooling} & Pool Size & 80\\
    & Strides & 2 \\
    \hline
    \multirow{2}{*}{Learning} &Rate & 0.0015\\
    &Decay & 3e-5\\
    \hline
    Bi-LSTM units && $2\times 500$\\
    Dropout rate && 0.15 \\
    Sequence padding& & 100\\
    Epochs & & 50\\
    Early stopping patience & & 5\\
    Batch size & &32\\
    Generated parameters& & 13M\\
    Average epoch time& & 2355ms\\
  \bottomrule
    \end{tabular}
    \end{table}

\subsection{Comparison Baselines}
We compare our results with the following supervised models. We refer to the main characteristics of each one in section \ref{sec:related}. CopyRE \cite{zeng2018extracting} and CopyMTL \cite{zeng2020copymtl} are based on the copy mechanism and used the same approach to generate word embeddings. Both evaluated their work on the Nyt and Webnlg datasets. GAME model \cite{cheng2022multi} used Spacy to generated word embeddings and reported their results on the Nyt dataset.

Other multiple relations classification models were not considered in the comparison due to their utilization of a different release of the Nyt dataset, such as \cite{li2020downstream} and ZSLRC \cite{gong2021zero}. We found that the used release is not commonly used in the literature.

\begin{table}[t]
\centering
\caption{Our models F1 score on the Nyt dataset compared with the baseline models.}
\label{tab:results-nyt}
\begin{tabular}{lcccc}
\toprule
 Model & GAME & CopyRE & CopyMTL & MRCA\\
\midrule
F1 & 77.1 & 87.0 & 86.9 & \textbf{96.65}$_{0.17}$\\
\bottomrule
\end{tabular}
\end{table}

\begin{table}[t]
\centering
\caption{Our models F1 score on the Webnlg dataset compared with the baseline models.}
\label{tab:results-webnlg}
\begin{tabular}{lccc}
\toprule
Model & CopyRE & CopyMTL & MRCA\\
\midrule
F1 & 75.1 & 79.7 & \textbf{93.35}$_{0.29}$\\
\bottomrule
\end{tabular}
\end{table}

\begin{table*}[h!]
\centering
\caption{The performance of our model's variants on the Webnlg dataset.}
\label{tab:ablation}
\begin{tabular}{llll}
\toprule
Model&Precision&Recall&F1\\
\midrule
MRCA&95.4$_{0.25}$&91.3$_{0.48}$&93.35$_{0.29}$\\
MRCA-Sigmoid-BCE&93.35$_{0.31}$&88.73$_{0.55}$&90.88$_{0.3}$\\
MRCA-Bert&94.5$_{0.2}$&89.9$_{0.49}$&92.15$_{0.26}$\\
MRCA-Bert-noLSTM&55.18$_{2.21}$&53.7$_{1.1}$&54.4$_{1.16}$\\
\bottomrule
\end{tabular}
\end{table*}

\subsection{Main Results and Analysis}
\begin{figure}[h!]
\centering
  \includegraphics[width=0.42\textwidth]{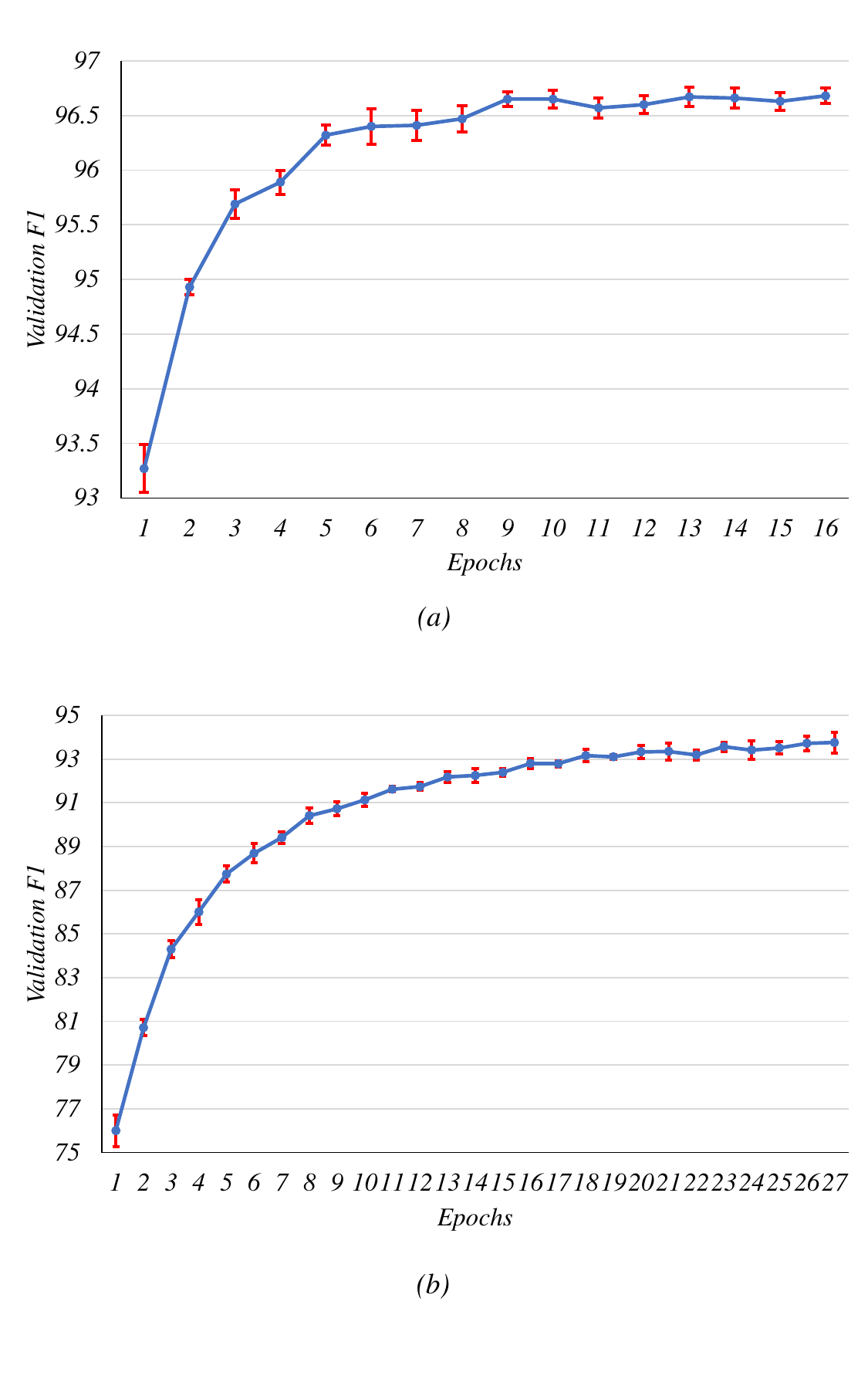}
  \caption{The validation F1 score during training for the evaluation datasets. (a) indicates the Nyt training performance. (b) indicates the Webnlg training performance.}
  \label{fig:train}
\end{figure}

We report our average F1 scores in Table \ref{tab:results-nyt} and Table \ref{tab:results-webnlg} for the Nyt and Webnlg datasets, respectively. Additionally, we visualize the training performance in Figure \ref{fig:train}. The results show superiority among the baseline models. We report the precision and recall scores in Table \ref{tab:ablation}. We highlight our results in the Webnlg dataset, as we find that relation predictions in that dataset is highly imbalanced due to the large number of predefined relations. Furthermore, the dataset has smaller training data. Nevertheless, the Webnlg's F1 score is close to the Nyt's score. Knowing that, the Nyt dataset has much smaller predefined relations and more training data, which indicates that our adaptation method supported achieving better predictions despite the imbalanced distribution of the binary labels.

\subsection{Ablation Study} \label{sec:ablation} To examine the effectiveness of our model's components, we evaluate the imbalanced predictions adaptation approach, and the text encoder adjustments. We design different variants of our model and perform training using the same main evaluation settings in Table \ref{tab:tuning}. Moreover, We report the average score of five runs and the standard deviation. We use the Webnlg dataset for the ablation study experiments. We report the performances in Table \ref{tab:ablation}, then we present the following analysis.

\textbf{Imbalanced Predictions Adaptation Effectiveness} To evaluate the contribution of our imbalanced predictions adaptation approach, we assess our model using different activation and loss functions. Specifically, we use the traditional \textit{sigmoid} activation function and the binary cross entropy loss function. We report this variant's performance in Table \ref{tab:ablation} with the name \textit{MRCA-Sigmoid-BCE}. The variant's F1 score is approximately 3\% less than our model's score, which is an average value between the precision scores difference and the recall scores difference. Noting that the recall gap is larger, which presents the first indication that the adaptation approach improved predicting negative labels.

\textbf{Encoder effectiveness} To evaluate our text encoder adjustments, we need to consider two sub-components in the assessments, that are the usage of Glove language model and the addition of the entity type vector to the embeddings. Thus, we test the following variants of our model. \textit{MRCA-Bert} is a variant the uses Bert PLM instead of Glove and \textit{MRCA-Bert-noLSTM} is a variant that uses Bert but with no LSTM layers. We use Bert's release\footnote{https://tfhub.dev/tensorflow/bert\_en\_cased\_L-12\_H-768\_A-12/4} with character case support since we added the same case feature in our implementation. In the former variant, there is a slight difference between the reported F1 score and our model' score, which demonstrates less contribution of the Glove employment in our overall performance. However, using Glove, our model still outperforms the Bert variant due to the better OOV terms support. Noting that Bert is known as a language model with contextual text representation support. Thus, the assumption is that, the LSTM layers would not affect Bert's performance. Nonetheless, in the second variant \textit{MRCA-Bert-noLSTM}, the performance is way worst. This result supports our claim that RC is not one of Bert's training objectives in section \ref{sec:encoder} because of the abstract usage of Bert. Furthermore, with a weak contextual representation in Bert, OOV words will split into non-meaningful tokens as described in the tokenization algorithm that is used in Bert \cite{song-etal-2021-fast}. This concludes the importance of using a language model with larger vocabulary.


\section{Conclusion} We propose MRCA, a multiple relations classification model that aims at improving the imbalanced predictions. Our light-design implementation leverages wider prediction range for negative labels and customize a remarkable loss function for the same purpose. Furthermore, text and entity features are utilized efficiently to improve the relations prediction. The experiments presented superiority among state-of-the-art models that reported the relation classification performance. Assessing our model's components showed that addressing the imbalanced predictions yields significant improvement in the relation classification task. Furthermore, representing sentences using language models with rich vocabularies provides performance enhancements in the relation classification task.

\section{Future Work and Limitations}
Although the relation classification task has limited applications as a single module, it has wider usages in the relation extraction task. Therefore, we see that our approach can be adopted to achieve new scores in several applications that utilize the relation classification task. Further improvements can be achieved when using NLP tools for lexical and syntactic text features. Additionally, it would be typical to advance our model to assign the predicted relation to the corresponding entities pair in the input. However, this approach cannot be considered as an ideal way for the relation or triple extraction task because errors in the entities tagging step would propagate to the relation classification task. Finally, our imbalanced predictions adaptation promises enhancements if used by similar tasks of imbalanced classes.

Our evaluation was limited by the small number of models that reported the relation classification performance. However, the results proved our model's superiority, denoted by the gap between our F1 score and the closest model.

\printbibliography 

@article{chen2020knowledge,
  title={Knowledge graph completion: A review},
  author={Chen, Zhe and Wang, Yuehan and Zhao, Bin and Cheng, Jing and Zhao, Xin and Duan, Zongtao},
  journal={Ieee Access},
  volume={8},
  pages={192435--192456},
  year={2020},
  publisher={IEEE}
}

@article{bundschus2008extraction,
  title={Extraction of semantic biomedical relations from text using conditional random fields},
  author={Bundschus, Markus and Dejori, Mathaeus and Stetter, Martin and Tresp, Volker and Kriegel, Hans-Peter},
  journal={BMC bioinformatics},
  volume={9},
  number={1},
  pages={1--14},
  year={2008},
  publisher={BioMed Central}
}

@article{vrandevcic2014wikidata,
  title={Wikidata: a free collaborative knowledgebase},
  author={Vrande{\v{c}}i{\'c}, Denny and Kr{\"o}tzsch, Markus},
  journal={Communications of the ACM},
  volume={57},
  number={10},
  pages={78--85},
  year={2014},
  publisher={ACM New York, NY, USA}
}

@inproceedings{riedel2010modeling,
  title={Modeling relations and their mentions without labeled text},
  author={Riedel, Sebastian and Yao, Limin and McCallum, Andrew},
  booktitle={ECML PKDD},
  pages={148--163},
  year={2010},
  organization={Springer}
}

@inproceedings{zhang2017tacred,
  author = {Zhang, Yuhao and Zhong, Victor and Chen, Danqi and Angeli, Gabor and Manning, Christopher D.},
  booktitle = {EMNLP},
  title = {Position-aware Attention and Supervised Data Improve Slot Filling},
  pages = {35--45},
  year = {2017}
}

@inproceedings{zeng2018extracting,
  title={Extracting relational facts by an end-to-end neural model with copy mechanism},
  author={Zeng, Xiangrong and Zeng, Daojian and He, Shizhu and Liu, Kang and Zhao, Jun},
  booktitle={ACL},
  pages={506--514},
  year={2018}
}

@article{yin2006efficient,
  title={Efficient classification across multiple database relations: A crossmine approach},
  author={Yin, Xiaoxin and Han, Jiawei and Yang, Jiong and Yu, Philip S},
  journal={IEEE Transactions on Knowledge and Data Engineering},
  volume={18},
  number={6},
  pages={770--783},
  year={2006},
  publisher={IEEE}
}

@inproceedings{wang2008re,
  title={A re-examination of dependency path kernels for relation extraction},
  author={Wang, Mengqiu},
  booktitle={Proceedings of the Third International Joint Conference on Natural Language Processing: Volume-II},
  year={2008}
}

@article{fundel2007relex,
  title={RelEx—Relation extraction using dependency parse trees},
  author={Fundel, Katrin and K{\"u}ffner, Robert and Zimmer, Ralf},
  journal={Bioinformatics},
  volume={23},
  number={3},
  pages={365--371},
  year={2007},
  publisher={Oxford University Press}
}

@article{klein2002fast,
  title={Fast exact inference with a factored model for natural language parsing},
  author={Klein, Dan and Manning, Christopher D},
  journal={Advances in neural information processing systems},
  volume={15},
  year={2002}
}

@article{gormley2015improved,
  title={Improved relation extraction with feature-rich compositional embedding models},
  author={Gormley, Matthew R and Yu, Mo and Dredze, Mark},
  journal={arXiv preprint arXiv:1505.02419},
  year={2015}
}

@article{zhang2018graph,
  title={Graph convolution over pruned dependency trees improves relation extraction},
  author={Zhang, Yuhao and Qi, Peng and Manning, Christopher D},
  journal={arXiv preprint arXiv:1809.10185},
  year={2018}
}

@inproceedings{hendrickx-etal-2010-semeval,
    title = "{S}em{E}val-2010 Task 8: Multi-Way Classification of Semantic Relations between Pairs of Nominals",
    author = "Hendrickx, Iris  and
      Kim, Su Nam  and
      Kozareva, Zornitsa  and
      Nakov, Preslav  and
      {\'O} S{\'e}aghdha, Diarmuid  and
      Pad{\'o}, Sebastian  and
      Pennacchiotti, Marco  and
      Romano, Lorenza  and
      Szpakowicz, Stan",
    booktitle = "Proceedings of the 5th International Workshop on Semantic Evaluation",
    month = jul,
    year = "2010",
    publisher = "Association for Computational Linguistics",
    pages = "33--38",
}

@inproceedings{wang2016relation,
  title={Relation classification via multi-level attention cnns},
  author={Wang, Linlin and Cao, Zhu and De Melo, Gerard and Liu, Zhiyuan},
  booktitle={Proceedings of the 54th Annual Meeting of the Association for Computational Linguistics (Volume 1: Long Papers)},
  pages={1298--1307},
  year={2016}
}

@inproceedings{zhou2016attention,
  title={Attention-based bidirectional long short-term memory networks for relation classification},
  author={Zhou, Peng and Shi, Wei and Tian, Jun and Qi, Zhenyu and Li, Bingchen and Hao, Hongwei and Xu, Bo},
  booktitle={Proceedings of the 54th annual meeting of the association for computational linguistics (volume 2: Short papers)},
  pages={207--212},
  year={2016}
}

@inproceedings{zeng2014relation,
  title={Relation classification via convolutional deep neural network},
  author={Zeng, Daojian and Liu, Kang and Lai, Siwei and Zhou, Guangyou and Zhao, Jun},
  booktitle={COLING},
  pages={2335--2344},
  year={2014}
}

@article{lee2019semantic,
  title={Semantic relation classification via bidirectional lstm networks with entity-aware attention using latent entity typing},
  author={Lee, Joohong and Seo, Sangwoo and Choi, Yong Suk},
  journal={Symmetry},
  volume={11},
  number={6},
  pages={785},
  year={2019},
  publisher={MDPI}
}

@inproceedings{ding2022relation,
  title={Relation Classification based on Selective Entity-Aware Attention},
  author={Ding, Haijie and Xu, Xiaolong},
  booktitle={CSCWD},
  pages={177--182},
  year={2022},
  organization={IEEE}
}

@article{mikolov2013efficient,
  title={Efficient estimation of word representations in vector space},
  author={Mikolov, Tomas and Chen, Kai and Corrado, Greg and Dean, Jeffrey},
  journal={arXiv preprint arXiv:1301.3781},
  year={2013}
}

@inproceedings{pennington2014glove,
  title={Glove: Global vectors for word representation},
  author={Pennington, Jeffrey and Socher, Richard and Manning, Christopher D},
  booktitle={EMNLP},
  pages={1532--1543},
  year={2014}
}

@article{devlin2018bert,
  title={Bert: Pre-training of deep bidirectional transformers for language understanding},
  author={Devlin, Jacob and Chang, Ming-Wei and Lee, Kenton and Toutanova, Kristina},
  journal={arXiv preprint arXiv:1810.04805},
  year={2018}
}

@inproceedings{wu2019enriching,
  title={Enriching pre-trained language model with entity information for relation classification},
  author={Wu, Shanchan and He, Yifan},
  booktitle={CIKM},
  pages={2361--2364},
  year={2019}
}

@inproceedings{baldini-soares-etal-2019-matching,
    title = "Matching the Blanks: Distributional Similarity for Relation Learning",
    author = "Baldini Soares, Livio  and
      FitzGerald, Nicholas  and
      Ling, Jeffrey  and
      Kwiatkowski, Tom",
    booktitle = "ACL",
    month = jul,
    year = "2019",
    publisher = "Association for Computational Linguistics",
    pages = "2895--2905",
}

@article{cohen2020relation,
  title={Relation classification as two-way span-prediction},
  author={Cohen, Amir DN and Rosenman, Shachar and Goldberg, Yoav},
  journal={arXiv preprint arXiv:2010.04829},
  year={2020}
}

@INPROCEEDINGS{9894216,
  author={Karaevli, Haluk Alper and Güngör, Tunga},
  booktitle={INISTA}, 
  title={Enhancing Relation Extraction by Using Shortest Dependency Paths Between Entities with Pre-trained Language Models}, 
  year={2022},
  volume={},
  number={},
  pages={1-7},
}

@article{qu2014senti,
  title={Senti-lssvm: Sentiment-oriented multi-relation extraction with latent structural svm},
  author={Qu, Lizhen and Zhang, Yi and Wang, Rui and Jiang, Lili and Gemulla, Rainer and Weikum, Gerhard},
  journal={Transactions of the Association for Computational Linguistics},
  volume={2},
  pages={155--168},
  year={2014},
  publisher={MIT Press One Rogers Street, Cambridge, MA 02142-1209, USA journals-info~…}
}

@inproceedings{zeng2020copymtl,
  title={Copymtl: Copy mechanism for joint extraction of entities and relations with multi-task learning},
  author={Zeng, Daojian and Zhang, Haoran and Liu, Qianying},
  booktitle={AAAI},
  volume={34},
  number={05},
  pages={9507--9514},
  year={2020}
}

@article{gu2016incorporating,
  title={Incorporating copying mechanism in sequence-to-sequence learning},
  author={Gu, Jiatao and Lu, Zhengdong and Li, Hang and Li, Victor OK},
  journal={arXiv preprint arXiv:1603.06393},
  year={2016}
}

@inproceedings{wang-etal-2019-extracting,
    title = "Extracting Multiple-Relations in One-Pass with Pre-Trained Transformers",
    author = "Wang, Haoyu  and
      Tan, Ming  and
      Yu, Mo  and
      Chang, Shiyu  and
      Wang, Dakuo  and
      Xu, Kun  and
      Guo, Xiaoxiao  and
      Potdar, Saloni",
    booktitle = "ACL",
    month = jul,
    year = "2019",
    publisher = "Association for Computational Linguistics",
    pages = "1371--1377",
}

@article{li2020downstream,
  title={Downstream model design of pre-trained language model for relation extraction task},
  author={Li, Cheng and Tian, Ye},
  journal={arXiv preprint arXiv:2004.03786},
  year={2020}
}

@article{paolini2021structured,
  title={Structured prediction as translation between augmented natural languages},
  author={Paolini, Giovanni and Athiwaratkun, Ben and Krone, Jason and Ma, Jie and Achille, Alessandro and Anubhai, Rishita and Santos, Cicero Nogueira dos and Xiang, Bing and Soatto, Stefano},
  journal={arXiv preprint arXiv:2101.05779},
  year={2021}
}

@article{cheng2022multi,
  title={Multi-Relation Extraction via A Global-Local Graph Convolutional Network},
  author={Cheng, Harry and Liao, Lizi and Hu, Linmei and Nie, Liqiang},
  journal={IEEE Transactions on Big Data},
  volume={8},
  number={6},
  pages={1716--1728},
  year={2022},
  publisher={IEEE}
}

@article{honnibal2017spacy,
  title={spaCy 2: Natural language understanding with Bloom embeddings, convolutional neural networks and incremental parsing},
  author={Honnibal, Matthew and Montani, Ines},
  journal={To appear},
  volume={7},
  number={1},
  pages={411--420},
  year={2017}
}

@inproceedings{gong2021zero,
  title={Zero-shot relation classification from side information},
  author={Gong, Jiaying and Eldardiry, Hoda},
  booktitle={CIKM},
  pages={576--585},
  year={2021}
}

@inproceedings{woodland2000effects,
  title={Effects of out of vocabulary words in spoken document retrieval},
  author={Woodland, Philip C and Johnson, Sue E and Jourlin, Pierre and Jones, K Sp{\"a}rck},
  booktitle={Proceedings of the 23rd annual international ACM SIGIR conference on Research and development in information retrieval},
  pages={372--374},
  year={2000}
}

@article{liu2019roberta,
  title={Roberta: A robustly optimized bert pretraining approach},
  author={Liu, Yinhan and Ott, Myle and Goyal, Naman and Du, Jingfei and Joshi, Mandar and Chen, Danqi and Levy, Omer and Lewis, Mike and Zettlemoyer, Luke and Stoyanov, Veselin},
  journal={arXiv preprint arXiv:1907.11692},
  year={2019}
}

@inproceedings{nayak2020domain,
  title={Domain adaptation challenges of BERT in tokenization and sub-word representations of out-of-vocabulary words},
  author={Nayak, Anmol and Timmapathini, Hariprasad and Ponnalagu, Karthikeyan and Venkoparao, Vijendran Gopalan},
  booktitle={Proceedings of the First Workshop on Insights from Negative Results in NLP},
  pages={1--5},
  year={2020}
}

@book{chollet2021deep,
  title={Deep learning with Python},
  author={Chollet, Francois},
  year={2021},
  publisher={Simon and Schuster},
pages={114, 60}
}

@inproceedings{li-etal-2020-dice,
    title = "Dice Loss for Data-imbalanced {NLP} Tasks",
    author = "Li, Xiaoya  and
      Sun, Xiaofei  and
      Meng, Yuxian  and
      Liang, Junjun  and
      Wu, Fei  and
      Li, Jiwei",
    booktitle = "ACL",
    month = jul,
    year = "2020",
    publisher = "Association for Computational Linguistics",
    pages = "465--476",
}

@article{huang2018robust,
  title={Robust liver vessel extraction using 3D U-Net with variant dice loss function},
  author={Huang, Qing and Sun, Jinfeng and Ding, Hui and Wang, Xiaodong and Wang, Guangzhi},
  journal={Computers in biology and medicine},
  volume={101},
  pages={153--162},
  year={2018},
  publisher={Elsevier}
}

@inproceedings{song-etal-2021-fast,
    title = "Fast {W}ord{P}iece Tokenization",
    author = "Song, Xinying  and
      Salcianu, Alex  and
      Song, Yang  and
      Dopson, Dave  and
      Zhou, Denny",
    booktitle = "EMNLP",
    month = nov,
    year = "2021",
    pages = "2089--2103",
}

\end{document}